\documentclass[letterpaper, 10 pt, conference]{ieeeconf}  

\IEEEoverridecommandlockouts                              
\renewcommand\keywords{\textit{\textbf{Keywords:}}}

\usepackage{graphicx} 
\usepackage{authblk}
\usepackage{breqn}

\title{\LARGE \bf
An approach to predictively securing critical cloud infrastructures through probabilistic modeling
}

\author{{Satvik Jain}\\
Dept. of Computer Engineering\\
Netaji Subhas Institute of Technology\\
University of Delhi, New Delhi, India\\
jainsatvik97@gmail.com
\and
{Arun Balaji Buduru}\\
Dept. of Computer Science\\
IIIT Delhi\\
New Delhi, India\\
arunb@iiitd.ac.in \\
\and 
{Anshuman Chhabra}\\
Dept. of Computer Science\\
University of California\\
Davis, USA\\
chhabra@ucdavis.edu}

\begin{document}

\maketitle
\thispagestyle{empty}
\pagestyle{empty}

\begin{abstract}
Cloud infrastructures are being increasingly utilized in critical infrastructures such as banking/finance, transportation and utility management. Sophistication and resources used in recent security breaches including those on critical infrastructures show that attackers are no longer limited by monetary/computational constraints. In fact, they may be aided by entities with large financial and human resources. Hence there is urgent need to develop predictive approaches for cyber defense to strengthen cloud infrastructures specifically utilized by critical infrastructures. Extensive research has been done in the past on applying techniques such as Game Theory, Machine Learning and Bayesian Networks among others for the predictive defense of critical infrastructures. However a major drawback of these approaches is that they do not incorporate probabilistic human behavior which limits their predictive ability. In this paper,  a stochastic approach is proposed to predict less secure states in critical cloud systems which might lead to potential security breaches. These less-secure states are deemed as `risky' states in our approach. 
Markov Decision Process (MDP) is used to accurately incorporate user behavior(s) as well as operational behavior of the cloud infrastructure through a set of features. The developed reward/cost mechanism is then used to select appropriate `actions' to identify risky states at future time steps by learning an optimal policy. Experimental results show that the proposed framework performs well in identifying  future `risky' states. Through this work we demonstrate the effectiveness of using probabilistic modeling (MDP) to predictively secure critical cloud infrastructures.\\

\end{abstract}

\section{INTRODUCTION}


Effective abstraction of services such as computing, networking and storage through efficient virtualization and distributed system technologies have ensured massive success of cloud applications. In cloud applications, since data is stored on servers, and the services are provided through software tools (such as web browsers), it ameliorates problem of scalability and flexibility in the fast-paced and dynamic real world environments.  With minimal investments, organizations and users can leverage services such as Infrastructure-as-a-Service (IaaS), Platform-as-a-Service (PaaS) and Software-as-a-Service (SaaS), among others. 

However, despite the obvious gains of utilizing cloud infrastructures in today's service ecosystem, there are many potential security issues that plague these systems. Securing cloud infrastructures is a tougher challenge compared to securing standalone systems due to the inherent `shared' nature of the cloud. Cloud systems not only have to deal with traditional network based attacks such as Denial of Service , Man in the Middle, Phishing attacks, but also counteract specifically personalized attacks. These include exploits targeting the use of shared technology \cite{kortchinsky2009cloudburst}, attacks made via insecure interfaces and APIs \cite{DBLP:journals/corr/ShahAS13}, cloud malware injection attacks \cite{5284165}, among others.


Since cloud systems are being increasingly adopted in critical infrastructures such as military, finance, utilities and transportation etc. the need of the hour is to build robust security frameworks. 
 In this paper, we propose a predictive security framework for critical cloud infrastructures at the sub-system level. Markov Decision Process (MDP) is used to model a sub-system and has representational features that can capture user behavior(s) and operational behavior of the sub-system.
The contributions of the proposed approach can be summarized into the following three areas: (1) A generalizable security framework with capability to continuously learn and predict future `risky' states, which in turn can be used to efficiently deploy and enforce security configurations, (2) Novel techniques to estimate the reward and utility of the states, and (3) Incorporation and utilization of probabilistic user behavior(s) which enables the system to be more robust and usable

In Section 2 we discuss the existing approaches to protect critical cloud infrastructures and their limitations. In Section 3 we outline the steps in our approach to predict 'risky' states in the subsystem of a cloud infrastructure. In Section 4 we detail the approach proposed in this work by discussing the implementation of each step of the general approach mentioned in the previous section . In Section 5, we describe the different experiments performed and results obtained by evaluating the accuracy of the proposed approach. In Section 6, we conclude by discussing the usefulness of our work along with directions for future work.

\section{Related Work}

Given the sophistication and resources used in some of the recent successful security breaches on critical cloud infrastructures, relying on reactive security techniques \cite{benameur2013cloud, ullah2013towards} is no longer sufficient. Providing sufficient deterrence-based security cover requires large amount of computational resources since these techniques must be actively deployed all the time. Further, they lack the ability to detect new generation threats which are targeted, persistent, stealthy, and unknown. Hence, there is a pressing need to rely on intelligent cyber security approaches having predictive ability to better protect critical cloud infrastructures.
%
%


Game theory is one of the most commonly used approaches to protect cyber infrastructures using predictive ability \cite{Shiva, 6636210, 4525297}. Game theoretic techniques are based on assumptions such as rationality of players, existence of a Nash Equilibrium and synchronization in actions of the attackers and defenders. However these constraints may not always hold true in real scenarios, which is one major pitfall of using Game Theoretic Techniques. Further, these techniques are not scalable with realistic sizes and complexity of the infrastructures. 



Recent work has also shown the effectiveness of applying machine learning and data mining techniques for cyber security \cite{4733933, article, TANG20113313, Kim:2008:HIF:1522798.1522832}. However in these techniques, models trained on a particular dataset become specific to mimicking the observations from history and therefore take time to adapt to unforeseen patterns. Keywhan Chung et al.\cite{7423125} in turn proposed a Q-Learning model which reacts automatically to the adversarial behavior of a suspicious user. However, the model does not provide any measure for quantifying the rewards of a successful attack or attack detection and relies on expert knowledge to enumerate the damage for taking a particular action. 



Probabilistic techniques used to protect cyber infrastructures include Bayesian networks \cite{6413615} and Markov Chains \cite{4299771, 5676911, 4539273}. Bayesian methods are able to deal with complex traffic distributions  by using probabilities obtained from historical data to calculate  probability of specific events. However it is very difficult to obtain prior distributions of a normal and an attack state. In the case of Markov chain models, the computations involved are relatively simple but there is apprehension in constructing the state profile of complex systems as all transition probabilities between possible states need to be calculated. Further, the predictive ability of these approaches is limited as they do not incorporate probabilistic human behavior in their attacker model.
Yau et. al \cite{7214175} introduced the possibility of using a combination of Bayesian networks and MDP for predictive security. However, the authors presented a very brief overview of their vision without illustrating how the MDP should be modeled and do not show any empirical evaluation to validate the proposed idea.

In an attempt to address the limitations posed by the predictive cyber security techniques mentioned above- most crucial being their lack of ability to capture probabilistic human behavior and in turn the system operational behavior, we have proposed an MDP based predictive approach for securing cloud infrastructures. To the best of our knowledge, an MDP based approach has never been used in the past work to model cloud infrastructure subsystems and further learn a policy to predict `risky' states at future time instants. Details of the proposed framework are presented in the following sections. 





\section{Proposed Approach}

In this paper, we aim to predict less secure states in a critical cloud sub-system which have  potential security breach risk. These less secure states of the cloud sub-system are termed as `risky' states in our approach. By identifying these future `risky' states prior to their occurence, the proposed model enables the cloud/security administrator to deploy the necessary security provisisons in time. We provide a predictive security approach by using Markov Decision Processes (MDP) to model the sub-system and capture the user and operational behavior(s). The MDP is solved to obtain an optimal policy which can lead the administrator to future `risky' states, given that currently the subsystem is in a `safe/non-risky' state. In contrast to previous techniques which do not incorporate user behavior/attacker modeling, our framework includes features representing user behavior. Since probabilistic human behavior affects the system operational behavior, it is imperative that predictive approaches should have some mechanism to facilitate the capturing and analysis of probabilistic human behavior accurately and efficiently. The policy learnt using our framework is based on information such as normal usage patterns, malicious behavior and subsystem performance metrics. Further, since the list of features is expandable, the proposed framework offers flexibility by allowing the  incorporation of more information in the future. Rather than proposing a model that completely  eliminates human intervention, our solution intends to assist administrators in better deploying and enforcing security. This allows for more robust security of the infrastructure. 

In this section, an outline of the approach with the steps involved in predicting risky states is provided. Implementation and specifics of each step in our model are given in Section 4. These steps are sequentially represented in Figure 1. Now we describe each step as shown in Figure 1:

%
%



\subsection{Gathering data from cloud infrastructure} 

\begin{figure}
\includegraphics[scale=0.35]{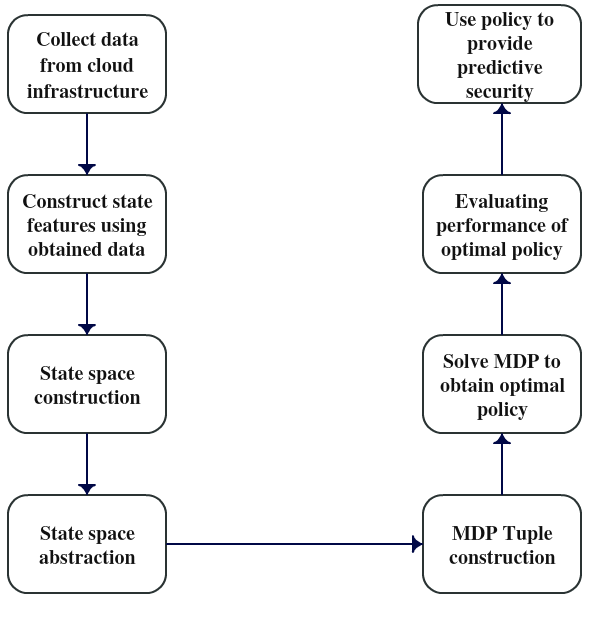}
\caption{Proposed approach as a flowchart}
\end{figure}

Critical cloud infrastructures can be thought of as a  composition of multiple subsystems or functional units, each serving a specific purpose. Each of these subsystems may be distributed over one or more virtual machines. To represent each sub-system as an MDP, data is collected to construct the features for the state space. Data for a specific time interval is collected for a cloud sub-system during which it would have both malicious as well as non-malicious incoming traffic.



\subsection{Constructing State Features}

Each MDP state is defined by a set of features/attributes. The data collected for a subsystem in the previous step is used to populate these features. The feature set can be divided into a finite number of categories with features in each category representative of a particular class of information. To model the cloud subsystem as an MDP we define the following three categories of features:   

\begin{enumerate}
\item \textit{Normal User Behavior:} This category contains features which characterize the normal traffic(non-malicious) incoming on the cloud subsystem. 
\item {Subsystem Performance:} This category contains features which characterize the performance of the Virtual Machine(s) hosting the subsystem.
\item \textit{Malicious Activity:} Contains features which can describe different types of  attacks such as DoS, CSRF, XSS and SQL injections on the subsystem.
\end{enumerate}

The above three categories are not exhaustive , however each of them has a variety of features providing sufficient information to define a state of the MDP. 


\subsection{State space construction}

To construct a finite state space, continuous data values under each feature need to be discretized. Different discretization techniques can be used such as Standard Binning, Fayyad and Irani's MDL method \cite{fayyad1993multi}, Class-attribute interdependence maximization (CAIM) \cite{kurgan2004caim} and Class-Attribute Contin-gency Coefficient (CACC) \cite{tsai2008discretization}






Selection of the method depends on the problem and data-set at hand. Once each feature has been discretized into values corresponding to each time step, the size of the generated state space for N features is evaluated as follows:

$|S|=n\textsubscript{1}*n\textsubscript{2}*n\textsubscript{3}*...n\textsubscript{N}$

In the above expression, $n\textsubscript{1},n\textsubscript{2},n\textsubscript{3}...n\textsubscript{N}$ are the number of unique values taken by each feature. 


\subsection{State Space Abstraction}

 Even after the feature set comprising of continuous data has been discretized in the previous step, the state space obtained is generally very large. State Space Abstraction is used to reduce the size of state-space of the MDP. Need for state space abstraction stems from the fact that most dynamic programming based algorithms that are used to solve the MDP have high overheads for  \emph{large} state spaces  \cite{littman1995complexity,papadimitriou1987complexity}. 
 Commonly used techniques for state space abstraction are Clustering based abstraction \cite{berkhin2006survey},Model-irrelevance abstraction \cite{li2006towards} and a\textsuperscript{*}-irrelevance abstraction \cite{li2006towards}. 
%
%



\subsection{MDP tuple construction}
A Markov Decision Process (MDP) is defined by tuple ($S$, $A$, $P$, $R$), where $S$ represents states, $A$ represents actions, $P$ : $S$x$A$x$S$ $\rightarrow [0, 1]$ represents a transition probability function and $R$ : $S$x$A\rightarrow R$ represents a reward function for each state-action pair. Actions are chosen to maximize an expected cumulative discounted reward:

\begin{equation}
V(s\textsubscript{n}) = E[\sum\limits_{n=0}^\infty \gamma\textsuperscript{n}R(s\textsubscript{n},a\textsubscript{n})]
\end{equation}

Here, $a\textsubscript{n} \in A$ is the action selected for current state $s\textsubscript{n} \in S$ and $\gamma \in (0, 1]$ is the discount factor. 

\subsection{Generating optimal policy}

Dynamic programming (DP) based algorithms such as value iteration \cite{vi} policy iteration \cite{pi} modified policy iteration \cite{mpi}, relative value iteration \cite{rvi} and Gauss Seidel value iteration \cite{gsvi} are used to solve the MDP and generate the optimal policy. The optimal policy aims to maximize the set of rewards or `value' that can be obtained at each state. 


\subsection{Evaluating optimal policy}
The evaluation of obatined optimal policy involves, first mapping the optimal policy from the abstracted state space to the original large state space and then determining it's accuracy. Accuracy of the obtained policy can be calculated as the percentage of instances where the policy correctly identified `risky' states. More details on how the optimal policy is evaluated are presented in Section 4-G.

\subsection{Identify future risky states }

If the accuracy achieved by the model at the previous step is satisfactory, it can be deployed to predict risky states at future time steps. The technique for identifying risky states of the critical cloud infrastructure subsystem at future time steps using the evaluated policy has been illustrated in detail in Section 4-H.  

\section{Implementation and Specifics}
%
%

We now describe the specifics for each step of the proposed solution outlined in the previous section. Each step of the previous section is covered here, with details concerning it's implementation, algorithm evaluation and design decisions.  

In this paper, to simulate the critical cloud infrastructure, a Banking application was deployed on Amazon Web Services using the Elastic Beanstalk platform. The Banking system consisted of three subsystems:Admin portal,Customer portal and Staff portal which were differentiated on basis of the user group accessing them. Cloud resources allocated to each subsystem included an EC2 instance (web server) and an RDS instance (database server). 


Since the aim of this work is to be able to predict potentially vulnerable states at the subsystem level of a critical cloud infrastructure, the admin portal was chosen as the subsystem for experimentation. The above choice of subsystem was not biased and the approach described in further steps can be replicated for any of the three subsystems. 

\subsection{Simulating traffic to obtain data}
%
%
The first step involves collecting data for state features of the MDP. Traffic simulations were performed on the cloud  subsystem in which both normal user traffic as well as malicious traffic was injected simultaneously. Now we describe the definition, configuration and tools used to simulate the two categories of traffic:

\begin{itemize}
\item \textit{Normal user traffic:} 

The term `normal user traffic' indicates the absence of any malicious activity in the form of attacks and comprises of only standard GET/POST http requests needed to perform various operations in the admin sub-system. \textit{Apache Jmeter} was used to simulate the normal traffic on the EC2 instance hosting the subsystem. 


\item \textit{Malicious traffic:} 

Three different types of flood attacks- SYN flood attacks, UDP flood attacks and the ICMP flood attacks were simulated on the EC2 instance hosting the subsystem. The \textit{Hping3} tool was used to generate these flood attacks. At the target server (that is, the EC2 instance hosting the subsystem) \textit{Snort} was deployed to detect the presence of a flood attack and generate corresponding alerts

%
%

\end{itemize}

Data for 300 seconds was collected from the server. This data was then used to generate features using the procedure described in the next step.

\subsection{Generate features using data}



We now discuss the description and evaluation of features in each of the three feature categories explained previously. Every feature value was evaluated for successive time steps of 1 second over the 300 second time window:
\begin{enumerate}
\item \textit{Normal user behavior:} 

\begin{enumerate}
\item \textbf{Number of http requests}: Total number of http-GET/POST requests on the EC2 server hosting the sub-system.
\item \textbf{Number of unique users}: Number of users logged in to the subsystem and performing various operations. 
\item \textbf{Requests-user distribution}: Distribution of http requests among the unique users logged into the subsystem.  It is the ratio of number of http requests and number of unique users. 
%
%
\item \textbf{Average bytes sent}: Calculates the average amount of data in bytes sent to the server through http requests. 
%
%
\end{enumerate}

\item \textit{Subsystem performance:} 


\begin{enumerate}
\item \textbf{Average latency}: 
Measures the average server latency during each 1 second time step. 

\item \textbf{Average response time}: Measures the average server response time during each 1 second time step. 

\end{enumerate}

\item \textit{Malicious Activity}: 
%
%

\begin{enumerate}

\item \textbf{DoS attack}: This feature describes whether or not any Denial-of-Service (DoS) attack attempts have been made on the host server. If any, it also differentiates between the three categories of Flood attacks simulated previously (SYN, UDP or ICMP based) and indicates the presence/absence of each during a time step. The presence of an attack is determined using the corresponding alert generated from \textit{Snort}.The reason for selecting DoS attacks specifically amongst other attack types in these experiments is because of the large-scale harm that DoS attacks are capable of causing. 


This feature is discrete-valued and takes 8 different values corresponding to all 8 occurence combinations of the three flood attacks.

\end{enumerate}



\end{enumerate}

It is important to note that apart from the last feature (DoS attacks) all the other features have continuous values. In order to define a state space having a finite number of states, each of these 6 features would need to be discretized. This is described in the next step.
\subsection{State space construction}
%
%
Standard binning technique  was used to convert the continuous range of values of the first six features into discrete integral values. This technique was chosen over the other techniques because of it's simplicity of evaluation, thus adding the least amount of computational overhead at this intermediate step. Further, this technique does not make any assumptions on the structure of data obtained post discretization nor introduces any bias like the other context based methods. 
The number of discrete values obtained for the 7 features were: 




\begin{itemize}
\item Number of http requests:5 (Range:0-50, Bin size:10)
\item Number of Unique Users: 5 (Range:0-50, Bin size: 10)
\item Requests-User Distribution:4 (Range:1-4, Bin size:0.75)
\item Average bytes sent:4 (Range:800-1300, Bin size:125)
\item Average latency:4 (Range:100-3500, Bin size: 850)
\item Average response time:4 (Range:0-8000,Bin size:2000) 
\item DoS attacks: 8
\end{itemize}

Based on the number of discrete values obtained for each feature, the size of state space is $5*5*4*4*4*4*8=51200$. 

\subsection{State space abstraction}

As evaluated in the last step, we have constructed a state space of 51200 states for the Markov Decision Process. However, as mentioned before, such a large state space is not practically solvable using most dynamic programming approaches which is why it becomes necessary to reduce the state space. 

In this paper, we opt for the clustering based abstraction method. Clustering based techniques perform unsupervised abstraction based on just the distribution of data points while most of the other state abstraction techniques such as model-irrelevance abstraction and a\textsuperscript{*}-irrelevance abstraction make some assumptions on the \emph{importance} of states to aggregate them. As we cannot label some states as either important or irrelevant in our model, it is appropriate to use a \emph{fair} algorithm that aggregates states without any bias. We use a number of clustering approaches for generating the abstracted state space in this step and find the one which gives the optimal performance. 

We use the following algorithms to perform the abstraction: K-Means using Euclidean distance metric (KME), K-Means using Mahalanobis distance metric (KMM) and Gaussian Mixture Models (GMM). To find the clustering algorithm best-suited to our use case amongst KME, KMM and GMM, we employ each of the algorithms and run through our proposed solution to find the highest accuracy achieved in the evaluation step (Section 4-G). The algorithm which gives the highest performance is chosen as the algorithm for carrying out the state space abstraction. The results and the details of the experiments regarding the choice of clustering algorithm are described in the Section 5.


%
%

\subsection{MDP Tuple construction}


Now we define the action set, reward matrix and transition matrices for the MDP representing a subsystem of the cloud banking infrastructure.

\subsubsection{\textbf{Action set}}: In our action set we define 2 actions:

\begin{itemize}
\item \textit{Remain in the same state}:  When this action is performed, there is a higher probability of remaining in the same state than of jumping to any different state. In the remaining text, this action is denoted by the integer `0'.
%
%
\item \textit{Jump to a different state}: When this action is performed, no  probability restriction is imposed on either jumping to a different state or remaining in the same state. In the remaining text, this action is denoted by the integer `1'.
\end{itemize}

The distinction between the above two actions becomes more clear with the description of the reward function and the transition function given below.

\subsubsection{\textbf{Reward function}}: This function returns the reward value for performing an action $a$ in state $s$. In our model, we define this reward value as follows: 
\begin{dmath}
R(s,a)= (w\textsubscript{1}*F\textsubscript{1})+(w\textsubscript{2}*F\textsubscript{2})+(w\textsubscript{3}*F\textsubscript{3})+(w\textsubscript{4}*F\textsubscript{4})+(w\textsubscript{5}*F\textsubscript{5})+(w\textsubscript{6}*F\textsubscript{6})+(w\textsubscript{7}*F\textsubscript{7})+(w\textsubscript{a}*R\textsubscript{a})
\end{dmath}
In the above expression, $w\textsubscript{1}$, $w\textsubscript{2}$, $w\textsubscript{3}$, $w\textsubscript{4}$, $w\textsubscript{5}$, $w\textsubscript{6}$, $w\textsubscript{7}$ are the weights given to each of the features proportional to their potential contribution in determining the `risk' associated with a state. 

From the description of features presented in Section 4-B, it can be seen that the feature category 3 (Malicious activity) is the most prominent indicator of potential security risk in a state, followed by features in Category 2 (Subsystem performance) and finally Category 1 (Normal user Behavior). Based on this logic the assigned weight values are $w\textsubscript{1} = 1000$, $w\textsubscript{2} = 1000$, $w\textsubscript{3} = 1000$, $w\textsubscript{4} = 1000$, $w\textsubscript{5} = 2000$, $w\textsubscript{6} = 2000$, $w\textsubscript{7} = 3000$.

$F\textsubscript{1}$, $F\textsubscript{2}$, $F\textsubscript{3}$, $F\textsubscript{4}$, $F\textsubscript{5}$, $F\textsubscript{6}$, $F\textsubscript{7}$ can take two values: either 0 or 1. If the value of a feature at the current state lies in the `safe range', corresponding value of $F\textsubscript{i}$ is 0, otherwise it is 1. The `safe range' values for each of the 7 features is taken as the first half of the range of values taken by that feature. This is based on the logic that as the integral value taken by a particular feature increases, the corresponding contribution of the feature in increasing the `risk' of a state also increases. Accordingly an unbiased approach to assign the `safe' values is to take the first half of the range of discretized values of a feature.



The last term in the reward metric expression, $(w\textsubscript{a}*R\textsubscript{a})$ takes into account the effect of action $a$ performed in a state. Thus, $w\textsubscript{a}$ is the weight and $R\textsubscript{a}$ is the reward term associated with action $a$. To define the term $R\textsubscript{a}$, we first define two terms: `Risk Metric' and `Risk Threshold'.

Risk metric $RM\textsubscript{s}$ gives the measure of `risk' associated with a particular state and is given by:
\begin{dmath}
RM\textsubscript{s} = w\textsubscript{1}*F\textsubscript{1}+w\textsubscript{2}*F\textsubscript{2}+w\textsubscript{3}*F\textsubscript{3}+w\textsubscript{4}*F\textsubscript{4}+w\textsubscript{5}*F\textsubscript{5}\\
+w\textsubscript{6}*F\textsubscript{6}
+w\textsubscript{7}*F\textsubscript{7}
\end{dmath}

`Risk threshold' ($R\textsubscript{th}$) is defined as: 
\begin{equation}
R\textsubscript{th} = \alpha*(w\textsubscript{1}+w\textsubscript{2}+w\textsubscript{3}+w\textsubscript{4}+w\textsubscript{5}+w\textsubscript{6}+w\textsubscript{7})
\end{equation}
Risk threshold is a fixed value such that if risk metric is greater than risk threshold, the state is termed as `risky' and if risk metric is less than risk threshold, the state is considered `non-risky'. Moreover, $\alpha$ can be any fixed value ranging between 0 to 1, depending on the criterion that the admin managing the cloud infrastructure sets to classify a state as `risky' or `non-risky'. A higher value of $\alpha$ implies a stricter criterion, and vice-versa. For our experiments, we set the value of $\alpha$ as 0.5, since dividing the risk metric into two ranges about the half-way mark introduces the least amount of bias towards either 'riskiness' or 'non-riskiness' of the state.  

Now that we have the risk metric and risk threshold, $R\textsubscript{a}$ is defined  using the following conditions:

\begin{itemize}
\item If $RM\textsubscript{s} > R\textsubscript{th}$:
\begin{enumerate}
\item If action $a$ is 0, then $R\textsubscript{a}$ is  +1
\item If action $a$ is 1, then $R\textsubscript{a}$ is -1
\end{enumerate}

\item If $RM\textsubscript{s} < R\textsubscript{th}$:
  \begin{enumerate}
  \item If action $a$ is 0, then $R\textsubscript{a}$ is -1 
  \item If action $a$ is 1, then $R\textsubscript{a}$ is +1
  \end{enumerate}
 \end{itemize}
  
  Since we want our optimal policy to identify the risky states, a positive reward of +1 is given in case the current state is `risky' and the action taken is `remain in the same state' or if the current  state is  `non-risky' and the action taken is `jump to a different state'. Apart from these two scenarios, a negative reward of -1 is given corresponding to the action taken. In order to make the range of the reward term $R\textsubscript{a}$ comparable to $RM\textsubscript{s}$ so that the outcome of the action taken has an observable effect on the value of $R(s,a)$, a weight $w\textsubscript{a}$ is multiplied with $R\textsubscript{a}$.Value of $w\textsubscript{a}$ is taken as average of the other seven weights ($w\textsubscript{1}$ to $w\textsubscript{7}$) associated with $RM\textsubscript{s}$.Average value has been taken to ensure fair contribution of the action in determining the reward value. This average value is approximately 1500.


The reward metric $R(s,a)$ is defined for a single state in the original state space. For abstract states (containing one or more states from the original state space) we calculate the average reward metric $R(s',a)$. If $s'$ is an abstract state containing the states $s1$,$s2$,$s3$...$sN$ from the original state space, then $R(s',a)$ is given by the following expression:
 \begin{equation}
 R(s',a)= (R(s1,a) + R(s2,a) + R(s3,a)+....+R(sN,a))/N
\end{equation}


Finally, a reward matrix of dimensions 1000 x 2 is generated. The 1000 rows correspond to the 1000 abstract states and the columns correspond to the actions `0' and `1'. The value of $R(s',a)$ derived above is used to generate the values that are inserted into the matrix.

\subsubsection{\textbf{Transition function:}} This function generates a probability value of reaching state $s'$ from state $s$ by performing an action $a$.

In order to calculate the transition probabilities, we use the data for the 300 time steps generated from traffic simulations on our cloud banking infrastructure. Each time step corresponds to a state in the original state space which in turn belongs to one of the abstracted states in the abstract state space. Hence, we generate an abstract state transition vector for the 300 time steps from the original data. 

 From the abstract state transitions data, the probability of transitioning from an abstract state $s$ to an abstract state $s'$ is given by:

\begin{equation}
P(s'|s) = N(s'|s) / N((s'' \neq s')|s)
\end{equation}

In the above equation, $N(s'|s)$ represents the number of transitions from $s$ to $s'$ in the abstract state transition vector. Here, $s$ is the abstract state at time step $t$, $s'$ is the abstract state at time step $t+1$. $N((s'' \neq s')|s)$ represents the number of transitions from $s$ to any abstract state $s''$ other than $s'$, from the state $s$, in the abstract state transition vector.

Thus, the transition function metric $P(s'|s,a)$ is calculated from $P(s'|s)$ as follows:

\begin{itemize}
\item If $a$ is 1 then $P(s'|s,a)=P(s'|s)$

\item If $a$ is 0 then:
\begin{itemize}
             \item If $s'= s$, then $P(s'|s,a) = t\textsubscript{s}$ where $t\textsubscript{s} \in (0.5,1]$ 
             \item If $s' \neq s$, then $P(s'|s,a)= (1-t\textsubscript s).*P(s'|s)$ 
\end{itemize}
\end{itemize}


$P(s'|s,a)$ when $s'= s$ is the case of self transitions, hence if the action taken is `0' (`remain in the same state') there is a very high stochastic probability of making the jump from $s$ to $s$ itself. Since this probability of self-transition should be directly dependent on the risk associated with the state $s$ (risk metric), we linearly map the risk metric of a particular state with the self-transition stochastic probability associated with it. The linear transformation is done such that $t\textsubscript{s}$ take values greater than 0.5 and less than or equal to 1. Thus the linear transformation will involve mapping from the $RM\textsubscript{s}$ range i.e. from $[min(RM\textsubscript{s}),max(RM\textsubscript{s})]$ to the range of $t\textsubscript{s}$, which is from (0.5,1]. Considering this linear transformation function to be $f(x)$, we can write:

\begin{equation}
f(x) = (\frac{x-min(x)}{max(x)-min(x)} * 0.49) + 0.51
\end{equation}

Using the above function, we get $f(RM\textsubscript{s}) = t\textsubscript{s}$. 



This function gives the self-transition probability for `risky' states as greater than 0.75, enforcing the logic that for these states, the action to `remain' in the same state is accompanied with a very high stochastic probability.

The non-self transition probability when $a$ is `0' is distributed from (1-t\textsubscript{s}) on the basis of $P(s'|s)$.

When $a$ is `1', the action taken is `Jump to a different state'. There is no imposed restriction and therefore all probabilities are calculated directly using $P(s'|s)$. 

Finally, two transition probability matrices of dimensions 1000 x 1000 corresponding to the actions `0' and `1' are created. The rows and columns correspond to the 1000 abstract states and the values in both matrices are filled from the above derived value of $P(s'|s,a)$. 


\subsection{Optimal policy generation}

Once the MDP is constructed, an optimal policy needs to be generated.  In our case, the optimal policy is a set of actions that can achieve the task of identifying risky states i.e. it should aim to `jump' from secure states and `remain' at risky states. Since we construct the MDP on the abstracted state space consisting of 1000 states, the optimal policy is a vector of length 1000, containing actions from the action set of the MDP. 

We intend to find the DP algorithm which gives us the `best' optimal policy for the MDP framework and also takes the minimum time to do so. Thus, in the results section (Section 5), we evaluate the DP algorithms not only on the basis of their accuracy but also on the time taken for them to solve the MDP. The intersection of these two performance metrics is chosen as the MDP solving algorithm for our framework. The results for these are obtained through experiments described in the next section.

\subsection{Evaluation of optimal policy}




The optimal policy for each of the 1000 states consists of either `jump' to the next state or `remain' in the current state. To map this logic back to the original state space, we award the same `optimal' action in the obtained policy from the abstracted (cluster) state to all the states in the original state space that belong to this cluster.

With the reverse mapping available, we can evaluate the optimal policy on both the abstracted state space and the original large state space. First, we find the `risky' states using the risk metric, $RM\textsubscript{s}$ (equation (3)) and risk threshold, $R\textsubscript{th}$ (equation (4)). If $RM\textsubscript{s}$ is greater than $R\textsubscript{th}$, the state is termed as 'risky'. 

With the states labeleled as either `risky' or `not-risky', we evaluate the performance of the optimal policy by calculating it's accuracy. This is done by calculating the ratio of states where the optimal policy gave a favourable outcome to the total number of states. Favourable outcomes are `remaining' at `risky' states  and `jumping' from `non-risky' states to identify the risky states. The numerical results of the accuracy obtained from the experiments performed are presented in Section 5. 


\subsection{Using the obtained optimal policy to predict risky states}

Once the optimal policy is obtained with a sufficient accuracy using the evaluation criterion mentioned in the previous step, it is used to identify future `risky' states. It is assumed that at the current time instant (taken as $t$=0), the subsystem is in a `non-risky' or `safe state'. At the current instant, the cloud infrastructure administrator would deploy the optimal policy starting from that safe state. As mentioned previously, the expected optimal policy is such that it generates the action `jump to a different state' in a non-risky state and `remain in the same state' in a risky state. Accordingly, once this optimal policy is deployed starting from the current safe state, a state transition tree is generated (according to the transition function defined in Section 4-E) extending into future time instants. This is shown in Figure 2.

\begin{figure}
\includegraphics[scale=0.39]{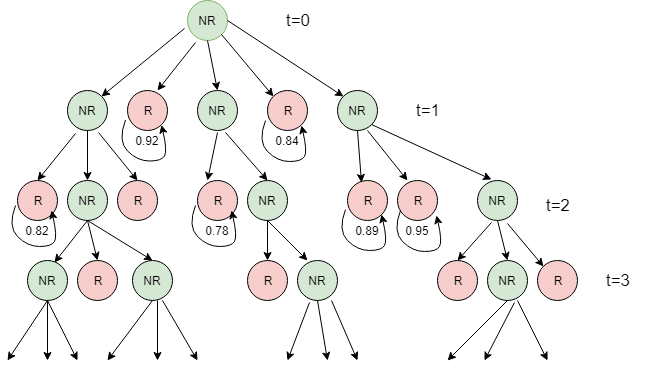}
\caption{Finding risky states using optimal policy}
\end{figure}

As seen from Figure 2, the risky states at any time step $t$ end up becoming the leaf nodes. This is because according to the transition function, defined in Section 4-E, the `remain in the same state' action taken by the optimal policy at risky states ensures a probability of at least 0.75 (refer to Section 4-E) of remaining in the same state and hence any further state transitions from that state would hold negligible probability. However, in the case of non-risky states, the action taken is `jump to a different state' and therefore that state becomes the root of a tree representing transitions from that state to other states. Therefore, from the tree generated in this manner, the risky states can be easily identified for any future time step as they will constitute all the leaf nodes. Furthermore, the probability of reaching a risky state can also be easily calculated by taking the product of the transition probabilities depicted on the branches of the tree, starting from the root node to the risky leaf node itself. 

Hence, at any given time instant when the subsystem is secure, the proposed approach can help the infrastructure administrator identify the risky states at future time steps as well as the probability of reaching those risky states. This information can be used to alert the admin so that necessary action can be taken to prevent the potential security breaches in time.

\section{Results and Analysis}

In this section, we present the results obtained by our framework using the experimental set-up described in in the previous section. We undertake three experiments as part of our proposed solution:

\subsection{Experiment 1: Choice of clustering algorithm for state abstraction}

We choose policy iteration (with discount factor, $\gamma$ as 0.1) as our MDP solving algorithm to obtain the optimal policy and then compute the accuracies on the original state space by varying the clustering algorithm for state space abstraction. We run experiments using K-Means with Euclidean distance metric (KME), K-Means with Mahalanobis distance metric (KMM) and Gaussian Mixture Models (GMM). We also vary the number of clusters between 250, 500, 750 and 1000 and see where we obtain the highest accuracy.

\begin{figure}
\caption{Finding optimal clustering algorithm}
\includegraphics[scale=0.30]{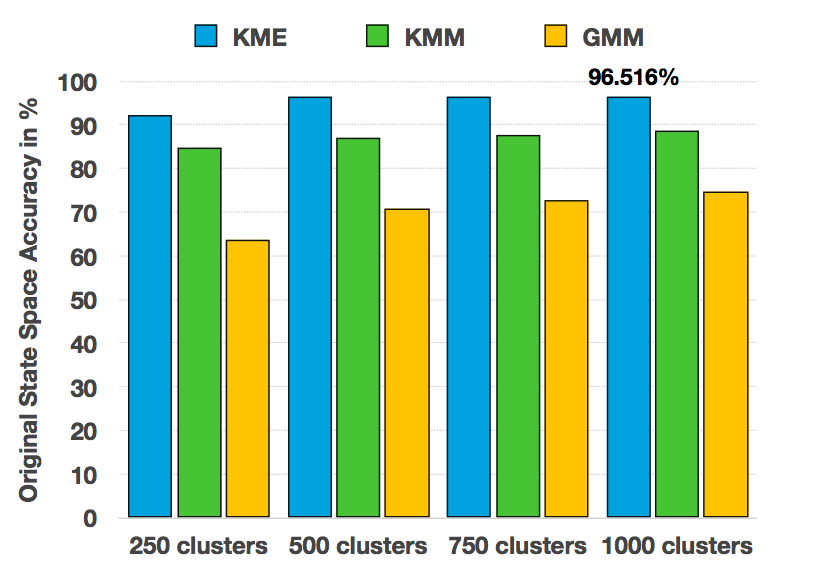}
\end{figure}

It can be seen in Figure 3, that KME for 1000 abstracted states obtains an accuracy of 96.516\% compared to KMM for 1000 clusters which obtains an accuracy of 88.775\% as well as GMM for 1000 clusters which achieves an accuracy of just 74.601\%. Here, policy iteration ($\gamma$ is 0.1) is used to obtain the optimal policy (although even value iteration, Gauss Seidel value iteration and modified policy iteration give the same optimal policies). We can also see in Figure 3 that for 1000 clusters, we obtain the highest original state space accuracy as compared to other cluster sizes in KME. However, to reaffirm the choice of 1000 clusters, we plot the mean square error (MSE) obtained while increasing cluster sizes. This is known as an elbow curve and is shown in Figure 4. It can be seen that the error tremendously decreases towards 1000 cluster size and does not decrease much after that. Therefore, we choose KME with 1000 abstracted states as our state space abstraction algorithm. 

\begin{figure}
\caption{The Elbow curve to find optimal cluster size}
\includegraphics[scale=0.30]{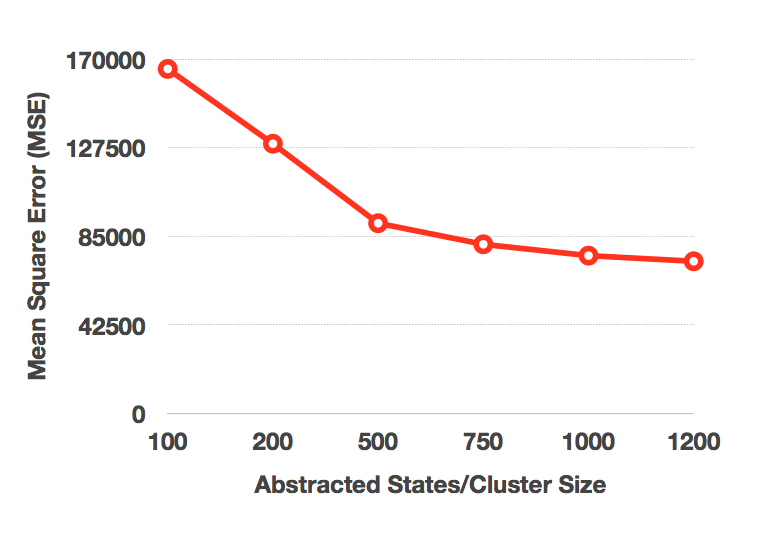}
\end{figure}

\subsection{Experiment 2: Finding best-suited MDP solving algorithm}

Through this experiment, we seek to find the optimal choice for the MDP solving algorithm giving the highest accuracy and minimal computation time. Here we choose KME with 1000 states as found in the previous experiment, as our clustering algorithm and compare modified policy iteration, policy iteration, value iteration, relative value iteration and Gauss Seidel value iteration to obtain the optimal policies and see which algorithm gives the highest original state space and abstracted state space accuracy. For all DP algorithms the value of discount factor ($\gamma$) is kept as 0.1.

We observe that modified policy iteration, policy iteration, value iteration and Gauss Seidel value iteration give the same optimal policy which achieves the highest original state accuracy of 96.516\% and highest abstracted space accuracy of 98.5\%. 
We then choose the best-suited algorithm out of these by finding the one that takes the minimum time to compute the optimal policy. As can be seen in Figure 5, modified policy iteration is the fastest algorithm out of all of the four with a completion time of 0.0335149 seconds. This is a much smaller value compared to even the next fastest algorithm, value iteration, which has a completion time of 0.81714 seconds. Therefore, modified policy iteration is chosen to be the algorithm to compute the optimal policy for our framework.


\subsection{Experiment 3: Finding optimal value of discount factor ($\gamma$)}

In the previous two experiments, $\gamma$ was chosen to be 0.1. However, we also need to experimentally find if a better choice of $\gamma$ exists that can yield higher accuracy for the original state space. In this experiment, we choose KME with 1000 states as the clustering algorithm and run modified policy iteration to solve the MDP, but with varying values of $\gamma$.

As can be seen in Figure 6, we vary $\gamma$ between 0.1 and 0.9 and find that on increasing $\gamma$, the accuracy for the original state space starts to decrease. The earlier obtained accuracy of 96.516\% for $\gamma$ set to 0.1, is the highest value of accuracy. Therefore, the optimal value for the discount factor is chosen to be 0.1.\\

Therefore, we have been able to find the best parameters and choice of algorithms for our framework. These are:

\begin{itemize}
\item K-Means with Euclidean metric for state space abstraction
\item Modified policy iteration for solving the MDP 
\item Discount factor ($\gamma$) set to 0.1
\end{itemize}

These settings give state-of-the-art results, with an accuracy of 98.5\% on abstracted state space and 96.516\% on the original large state space.

Since the obtained optimal policy is highly accurate (performance accuracy is 96.516\%) with respect to the expected optimal policy of `remaining' in risky states and `jumping' to different states while in a non-risky state, it can be confidently used to identify future risky states along with their probabilities using the approach mentioned in Section 4-H.

\begin{figure}
\caption{Completion times for MDP solving algorithms}
\includegraphics[scale=0.30]{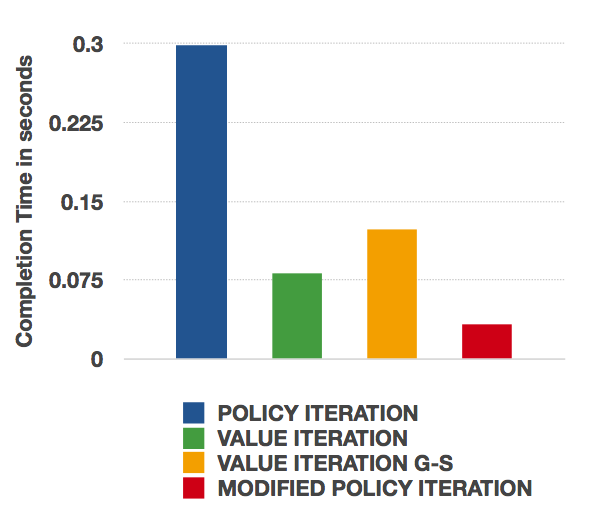}
\end{figure}

\section{Conclusion and Future Work}


In this paper, an approach for predictively securing critical cloud infrastructures is developed and evaluated. The framework utilizes MDP at the sub-system level to capture probabilistic user behavior and operational behavior of the sub-system through a set of features which define the MDP state space. 
Further, a suitable reward function is defined through which the learnt optimal policy is able to identify future `risky' states. These `risky' states can lead to potential security breaches. The step-wise procedure of the proposed approach have been detailed in the aforementioned sections. Various experimental evaluations are performed in order to maximize the prediction accuracy of the generated policy. These include:  (1) comparing different clustering algorithms for state-space abstraction; (2) empirically determining that modified policy iteration provides the highest accuracy and has least convergence time among other DP techniques. The resulting framework is designed based on the above evaluations and achieves an accuracy of 96.516\% in identifying future `risky' states . This reflects the effectiveness of using probabilistic modeling through MDP to predictively secure critical cloud infrastructures. 

\begin{figure}
\caption{Finding best value of $\gamma$}
\includegraphics[scale=0.30]{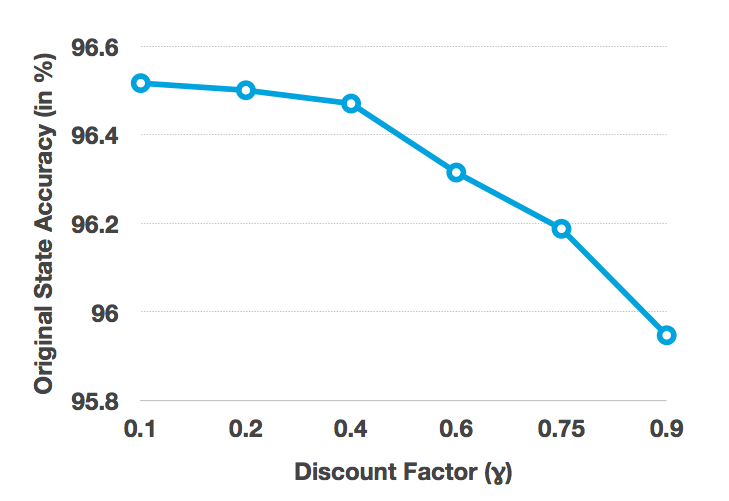}
\end{figure}

  

Future work primarily aims at expanding the feature set especially in the `malicious activity' category by including more sophisticated and varied attack models to ensure further robustness of the security framework. In addition, for experimental feasibility, traffic simulations on the cloud infrastructure were performed at a relatively small-scale. The step-wise approach of our proposed framework will be undertaken for more large scale cloud systems. Furthermore, the proposed MDP framework for predicting security breaches at the subsystem level of the cloud infrastructure can be extended by using it in combination with Bayesian Networks to predict security breaches at the system-wide level. Each subsystem can send the predicted risky states to the system-wide Bayesian Network, either directly or indirectly. Utilization of Bayesian Network has much lesser overhead and is better equipped to monitor all the events which occur across subsystems, and thus predict system-wide security breaches.



\bibliographystyle{IEEEtran}
\bibliography{main}

\begin{thebibliography}{10}
\providecommand{\url}[1]{#1}
\csname url@samestyle\endcsname
\providecommand{\newblock}{\relax}
\providecommand{\bibinfo}[2]{#2}
\providecommand{\BIBentrySTDinterwordspacing}{\spaceskip=0pt\relax}
\providecommand{\BIBentryALTinterwordstretchfactor}{4}
\providecommand{\BIBentryALTinterwordspacing}{\spaceskip=\fontdimen2\font plus
\BIBentryALTinterwordstretchfactor\fontdimen3\font minus
  \fontdimen4\font\relax}
\providecommand{\BIBforeignlanguage}[2]{{%
\expandafter\ifx\csname l@#1\endcsname\relax
\typeout{** WARNING: IEEEtran.bst: No hyphenation pattern has been}%
\typeout{** loaded for the language `#1'. Using the pattern for}%
\typeout{** the default language instead.}%
\else
\language=\csname l@#1\endcsname
\fi
#2}}
\providecommand{\BIBdecl}{\relax}
\BIBdecl

\bibitem{kortchinsky2009cloudburst}
K.~Kortchinsky, ``Cloudburst: Hacking 3d (and breaking out of vm-ware). url:
  https://www. blackhat. com/presentations/bh-usa-09/kortchinsky,''
  \emph{BHUSA09-Kortchinsky-Cloudburst-SLIDES. pdf (vid. p{\'a}g. 13)}, 2009.

\bibitem{DBLP:journals/corr/ShahAS13}
\BIBentryALTinterwordspacing
H.~Shah, S.~S. Anandane, and Shrikanth, ``Security issues on cloud computing,''
  \emph{CoRR}, vol. abs/1308.5996, 2013. [Online]. Available:
  \url{http://arxiv.org/abs/1308.5996}
\BIBentrySTDinterwordspacing

\bibitem{5284165}
M.~Jensen, J.~Schwenk, N.~Gruschka, and L.~L. Iacono, ``On technical security
  issues in cloud computing,'' in \emph{2009 IEEE International Conference on
  Cloud Computing}, Sept 2009, pp. 109--116.

\bibitem{benameur2013cloud}
A.~Benameur, N.~S. Evans, and M.~C. Elder, ``Cloud resiliency and security via
  diversified replica execution and monitoring,'' in \emph{Resilient Control
  Systems (ISRCS), 2013 6th International Symposium on}.\hskip 1em plus 0.5em
  minus 0.4em\relax IEEE, 2013, pp. 150--155.

\bibitem{ullah2013towards}
K.~W. Ullah, A.~S. Ahmed, and J.~Ylitalo, ``Towards building an automated
  security compliance tool for the cloud,'' in \emph{Trust, Security and
  Privacy in Computing and Communications (TrustCom), 2013 12th IEEE
  International Conference on}.\hskip 1em plus 0.5em minus 0.4em\relax IEEE,
  2013, pp. 1587--1593.

\bibitem{Shiva}
\BIBentryALTinterwordspacing
S.~Shiva, S.~Roy, and D.~Dasgupta, ``Game theory for cyber security,'' in
  \emph{Proceedings of the Sixth Annual Workshop on Cyber Security and
  Information Intelligence Research}, ser. CSIIRW '10.\hskip 1em plus 0.5em
  minus 0.4em\relax New York, NY, USA: ACM, 2010, pp. 34:1--34:4. [Online].
  Available: \url{http://doi.acm.org/10.1145/1852666.1852704}
\BIBentrySTDinterwordspacing

\bibitem{6636210}
A.~Pătraşcu and E.~Simion, ``Game theory in cyber security defence,'' in
  \emph{Proceedings of the International Conference on ELECTRONICS, COMPUTERS
  and ARTIFICIAL INTELLIGENCE - ECAI-2013}, June 2013, pp. 1--6.

\bibitem{4525297}
W.~Jiang, Z.~h.~Tian, H.~l.~Zhang, and X.~f.~Song, ``A stochastic game
  theoretic approach to attack prediction and optimal active defense strategy
  decision,'' in \emph{2008 IEEE International Conference on Networking,
  Sensing and Control}, April 2008, pp. 648--653.

\bibitem{4733933}
O.~Thonnard and M.~Dacier, ``Actionable knowledge discovery for threats
  intelligence support using a multi-dimensional data mining methodology,'' in
  \emph{2008 IEEE International Conference on Data Mining Workshops}, Dec 2008,
  pp. 154--163.

\bibitem{article}
H.~Farhadi, M.~Amir~Haeri, and M.~Khansari, ``Alert correlation and prediction
  using data mining and hmm,'' vol.~3, pp. 77--102, 01 2011.

\bibitem{TANG20113313}
\BIBentryALTinterwordspacing
C.~Tang, Y.~Xie, B.~Qiang, X.~Wang, and R.~Zhang, ``Security situation
  prediction based on dynamic bp neural with covariance,'' \emph{Procedia
  Engineering}, vol.~15, pp. 3313 -- 3317, 2011, cEIS 2011. [Online].
  Available:
  \url{http://www.sciencedirect.com/science/article/pii/S1877705811021229}
\BIBentrySTDinterwordspacing

\bibitem{Kim:2008:HIF:1522798.1522832}
\BIBentryALTinterwordspacing
S.~Kim, S.-j. Shin, H.~Kim, K.~H. Kwon, and Y.~Han, ``Hybrid intrusion
  forecasting framework for early warning system,'' \emph{IEICE - Trans. Inf.
  Syst.}, vol. E91-D, no.~5, pp. 1234--1241, May 2008. [Online]. Available:
  \url{http://dx.doi.org/10.1093/ietisy/e91-d.5.1234}
\BIBentrySTDinterwordspacing

\bibitem{7423125}
K.~Chung, C.~A. Kamhoua, K.~A. Kwiat, Z.~T. Kalbarczyk, and R.~K. Iyer, ``Game
  theory with learning for cyber security monitoring,'' in \emph{2016 IEEE 17th
  International Symposium on High Assurance Systems Engineering (HASE)}, Jan
  2016, pp. 1--8.

\bibitem{6413615}
J.~Wu, L.~Yin, and Y.~Guo, ``Cyber attacks prediction model based on bayesian
  network,'' in \emph{2012 IEEE 18th International Conference on Parallel and
  Distributed Systems}, Dec 2012, pp. 730--731.

\bibitem{4299771}
D.~H. Kim, T.~Lee, S.~O.~D. Jung, H.~P. In, and H.~J. Lee, ``Cyber threat trend
  analysis model using hmm,'' in \emph{Third International Symposium on
  Information Assurance and Security}, Aug 2007, pp. 177--182.

\bibitem{5676911}
D.~Man, Y.~Wang, W.~Yang, and W.~Wang, ``A combined prediction method for
  network security situation,'' in \emph{2010 International Conference on
  Computational Intelligence and Software Engineering}, Dec 2010, pp. 1--4.

\bibitem{4539273}
D.~S. Fava, S.~R. Byers, and S.~J. Yang, ``Projecting cyberattacks through
  variable-length markov models,'' \emph{IEEE Transactions on Information
  Forensics and Security}, vol.~3, no.~3, pp. 359--369, Sept 2008.

\bibitem{7214175}
S.~S. Yau, A.~B. Buduru, and V.~Nagaraja, ``Protecting critical cloud
  infrastructures with predictive capability,'' in \emph{2015 IEEE 8th
  International Conference on Cloud Computing}, June 2015, pp. 1119--1124.

\bibitem{fayyad1993multi}
U.~Fayyad and K.~Irani, ``Multi-interval discretization of continuous-valued
  attributes for classification learning,'' 1993.

\bibitem{kurgan2004caim}
L.~A. Kurgan and K.~J. Cios, ``Caim discretization algorithm,'' \emph{IEEE
  transactions on Knowledge and Data Engineering}, vol.~16, no.~2, pp.
  145--153, 2004.

\bibitem{tsai2008discretization}
C.-J. Tsai, C.-I. Lee, and W.-P. Yang, ``A discretization algorithm based on
  class-attribute contingency coefficient,'' \emph{Information Sciences}, vol.
  178, no.~3, pp. 714--731, 2008.

\bibitem{littman1995complexity}
M.~L. Littman, T.~L. Dean, and L.~P. Kaelbling, ``On the complexity of solving
  markov decision problems,'' in \emph{Proceedings of the Eleventh conference
  on Uncertainty in artificial intelligence}.\hskip 1em plus 0.5em minus
  0.4em\relax Morgan Kaufmann Publishers Inc., 1995, pp. 394--402.

\bibitem{papadimitriou1987complexity}
C.~H. Papadimitriou and J.~N. Tsitsiklis, ``The complexity of markov decision
  processes,'' \emph{Mathematics of operations research}, vol.~12, no.~3, pp.
  441--450, 1987.

\bibitem{berkhin2006survey}
P.~Berkhin, ``A survey of clustering data mining techniques,'' in
  \emph{Grouping multidimensional data}.\hskip 1em plus 0.5em minus 0.4em\relax
  Springer, 2006, pp. 25--71.

\bibitem{li2006towards}
L.~Li, T.~J. Walsh, and M.~L. Littman, ``Towards a unified theory of state
  abstraction for mdps.''

\bibitem{vi}
\BIBentryALTinterwordspacing
R.~Bellman, \emph{Dynamic Programming}, ser. Dover Books on Computer
  Science.\hskip 1em plus 0.5em minus 0.4em\relax Dover Publications, 2013.
  [Online]. Available: \url{https://books.google.co.in/books?id=CG7CAgAAQBAJ}
\BIBentrySTDinterwordspacing

\bibitem{pi}
\BIBentryALTinterwordspacing
L.~G. Telser, ``Dynamic programming and markov processes. ronald a. howard,''
  \emph{Journal of Political Economy}, vol.~69, no.~3, pp. 296--297, 1961.
  [Online]. Available: \url{https://doi.org/10.1086/258477}
\BIBentrySTDinterwordspacing

\bibitem{mpi}
M.~L. Puterman and M.~C. Shin, ``Modified policy iteration algorithms for
  discounted markov decision problems,'' \emph{Management Science}, vol.~24,
  no.~11, pp. 1127--1137, 1978.

\bibitem{rvi}
\BIBentryALTinterwordspacing
D.~White, ``Dynamic programming, markov chains, and the method of successive
  approximations,'' \emph{Journal of Mathematical Analysis and Applications},
  vol.~6, no.~3, pp. 373 -- 376, 1963. [Online]. Available:
  \url{http://www.sciencedirect.com/science/article/pii/0022247X63900179}
\BIBentrySTDinterwordspacing

\bibitem{gsvi}
\BIBentryALTinterwordspacing
N.~A.~J. Hastings, ``Optimization of discounted markov decision problems,''
  \emph{Journal of the Operational Research Society}, vol.~20, no.~4, pp.
  499--500, 1969. [Online]. Available:
  \url{https://doi.org/10.1057/jors.1969.112}
\BIBentrySTDinterwordspacing

\end{thebibliography}


\end{document}